\documentclass{article}

     \PassOptionsToPackage{numbers, compress}{natbib}

\usepackage[final]{neurips_2019}

\usepackage[utf8]{inputenc} 
\usepackage[T1]{fontenc}    
\usepackage{hyperref}       
\usepackage{url}            
\usepackage{booktabs}       
\usepackage{amsfonts}       
\usepackage{nicefrac}       
\usepackage{microtype}      

\usepackage{graphicx}
\usepackage{algorithm}
\usepackage{algorithmic}
\usepackage{amsthm}
\usepackage{amssymb}
\usepackage{multicol}
\usepackage{times}

\usepackage{amsmath}
\newtheorem{thm}{Theorem}
\newtheorem{lem}{Lemma}

\newtheorem{definition}{Definition}

\title{Online learning with Corrupted context: Corrupted Contextual Bandits}

\begin{document}
\author{%
   Djallel Bouneffouf  \\
   IBM Research  \\
   \texttt{djallel.bouneffouf@ibm.com} \\
}
\maketitle

\begin{abstract}
We consider a novel variant of the contextual bandit problem (i.e., the multi-armed bandit with side-information, or context, available to a decision-maker) where the context used at each decision may be corrupted ("useless context"). This new problem is motivated by certain on-line settings including clinical trial and ad recommendation applications. In order to address the corrupted-context setting, we propose to combine the standard contextual bandit approach with a classical multi-armed bandit mechanism. Unlike standard contextual bandit methods, we are able to learn from all iteration, even those with corrupted context, by improving the computing of the expectation for each arm. Promising empirical results are obtained on several real-life datasets.
\end{abstract}

\section{Introduction}
Sequential decison making \cite{BouneffoufBG13,ChoromanskaCKLR19,RiemerKBF19,LinC0RR20,lin2020online,lin2020unified} is a common problem in many practical applications  where the agent must choose the best action to perform at each iteration in order to maximize the cumulative reward over some period of time ~\cite {surveyDB,T33}. One of the key challenges is achieve a good trade-off between the exploration of new actions and the exploitation of known actions \cite{BouneffoufBG12}.  This exploration vs exploitation trade-off in sequential decision making problems is often formulated as  the {\em multi-armed bandit (MAB)} problem: given a set of bandit ``arms'' (actions),  each associated with a fixed but unknown reward probability distribution ~\cite {LR85,UCB,Bouneffouf0SW19,LinBCR18,DB2019,BalakrishnanBMR19ibm,BouneffoufLUFA14,RLbd2018},  an agent selects an arm to play at each iteration, and receives a reward,  drawn according to the selected arm's distribution, independently from the previous actions. 

A particularly useful version of MAB is the {\em contextual multi-armed bandit (CMAB)}, or simply the {\em contextual bandit} problem, where at  each iteration,  before choosing an arm, the agent observes an $N$-dimensional {\em context}, or {\em feature vector}.
Over time, the goal is to learn the relationship between the context vectors and the rewards, in order to make better prediction which action to choose   given the  context \cite{AuerC98,AuerCFS02,BalakrishnanBMR18}.
For example, the  contextual bandit approach is commonly used in various practical sequential decision problems with side information (context), from clinical trials \cite{villar2015multi} to recommender system \cite{MaryGP15,Bouneffouf16,aaai0G20}, where the patient's information (medical history, etc.) or an online user's profile provide a context for making a better decision about  the treatment to propose or an ad to show, and the   reward represents  the outcome  of the selected action, such as, for example, success or failure of a particular treatment option.
 
 In this paper, we consider a new problem setting, referred to as contextual bandit with corrupted context, where the agent may not always observe the true context. This setting is motivated by several real-life applications. 
 For instance, in online advertisement, the user in front of the screen is not the usual user who login (could be his brother), so the user profile (context) that the recommender system is using to recommend is not the right one. Another example, in medical decision-making settings, the doctor order some blood test but the result are wrong due to a problem in the machine. So the doctor is using a corrupted context (blood test results) to make decisions.
  
The corrupted contextual bandit framework proposed here aims to capture the situations described above, and provides an approach to always exploiting the current interaction in order to improve future decisions. More specifically, we will combine contextual bandit algorithm with the classical multi-armed bandit: the MAB allows to learn reward estimate with and without the observation of the right context, while the contextual bandit makes use of the right context information when it is available. We demonstrate on several real-life datasets that the proposed approach consistently outperforms the standard contextual bandit approach.

Overall, the main contributions of this paper include (1) a new formulation of a bandit problem, motivated by practical applications, (2) a new algorithm for stationary settings of the corrupted-context contextual bandit problem, (3) theoretical analysis of the regret bound for the algorithm, and (4) empirical evaluation demonstrating advantages of the proposed methods over a range of datasets and parameter settings.

This paper is organized as follows. Section \ref{sec:related} reviews related works. Section \ref{background} introduces some background concepts. Section \ref{sec:statement} introduces the contextual bandit model with corrupted context, and the proposed algorithm. Experimental evaluation on several datasets, for varying parameter settings, is presented in Section \ref{sec:experimental}. Finally, the last section concludes the paper and points out possible directions for future works.
\section{Related Work}
\label{sec:related}
The multi-armed bandit problem has been extensively studied. Different solutions have been proposed using a stochastic formulation ~\cite {LR85,UCB,BouneffoufF16} and a Bayesian formulation ~\cite {AgrawalG12}; however, these approaches did not take into account the context. In LINUCB ~\cite{Li2010}, Neural Bandit \cite{AllesiardoFB14} and in Contextual Thompson Sampling (CTS)~\cite{AgrawalG13}, a linear dependency is assumed between the expected reward of an action and its context; the representation space is modeled using a set of linear predictors.
However, the observed context is assumed to be the true one, unlike in this paper. Authors in \cite{YadkoriPS12} studied a sparse variant of stochastic linear bandits, where only a relatively small (unknown) subset of features is relevant to a multivariate function optimization. For high-dimensional bandits, it presents an application to the problem of optimizing a function that depends on many features, where only a small, initially unknown subset of features is relevant. Similarly, \cite{CarpentierM12} also considered the high-dimensional stochastic linear bandits with sparsity, combining the ideas from compressed sensing and bandit theory.

In \cite{gajanecorrupt} authors study a variant of the stochastic multi-armed bandit (MAB) problem in which the rewards are corrupted. In this framework, motivated by privacy preserving in online recommender systems, the goal is to maximize the sum of the (unobserved) rewards, based on the observation of transformation of these rewards through a stochastic corruption process with known parameters. However, this work was done in the classical multi-armed bandit setting with no context.

Different paper assume some constrain on the features, for instance, authors in \cite{BouneffoufRCcF17} consider a novel formulation of the contextual bandit problem when there are constraints on the context, i.e., where only a limited number of features can be accessed by the learner at each iteration.

In \cite{YunNMS17, SharmaZABMV20}, the authors study the contextual multi-armed bandit problems assuming uncertainty on features. Specifically, they assume that the learner is observing a noisy context vector where random noise is independently drawn from some distribution. 

Note that {\em none of the previous approaches addresses the problem of the contextual bandit setting when the context could be useless due to some corruption}, which is the main focus of this work. 
\section{Key Notion}
This section focuses on introducing the key notions used in this paper.

\section{Background}
\label{background}
This section  introduces some background concepts  our approach builds upon, such as contextual bandit, and Thompson Sampling.
 \subsection{The contextual bandit problem}
Following \cite{langford2008epoch}, this problem is defined as follows.
At each time point (iteration) $t \in \{1,...,T\}$, a player is presented with a {\em context} ({\em feature vector}) $\textbf{c}(t) \in \mathbf{R}^N$
  before choosing an arm $k  \in A = \{ 1,...,K\} $.
We will denote by
  $C=\{C_1,...,C_N\}$  the set of features (variables) defining the context.
Let ${\bf r} = (r_{1}(t),...,$ $r_{K}(t))$ denote  a reward vector, where $r_k(t) \in [0,1]$ is a reward at time $t$  associated with the arm $k\in A$.
Herein, we will primarily focus on the Bernoulli bandit with binary reward, i.e. $r_k(t) \in \{0,1\}$.
Let $\pi: C \rightarrow A$ denote a policy.  Also, $D_{c,r}$ denotes a joint distribution  $({\bf c},{\bf r})$.
We will assume that the expected reward is a  linear function of the context, i.e.
$E[r_k(t)|\textbf{c}(t)] $ $= \mu_k^T \textbf{c}(t)$,
where $\mu_k$ is an unknown weight vector (to be learned from the data) associated with the arm $k$. \\


\section{Problem Setting}
\label{sec:statement}
In this section, we define a new type of a bandit problem, the {\em contextual bandit with corrupted context (CBCC)};
 our approach will be based on the Thompson Sampling \cite{AgrawalG12}.

\subsection{Contextual Bandit with Corrupted Context (CBCC)}
Algorithm \ref{alg:CBP} presents at a high-level the CBCC setting, where $\psi: c \rightarrow \hat{c}$ denotes a corrupting function and $\hat{c}$ the corrupted context, $p_{\psi}$ is the probability that the context is corrupted by the function $\psi$. Note, in this setting, we assume that the function $\psi$ is unknown to the player and could not be recovered.
\begin{algorithm}[H]
	\caption{ The CBCC Problem Setting}
	\label{alg:CBP}
	\begin{algorithmic}[1]
	 \STATE {\bfseries }\textbf{Foreach} $t = 1, 2, . . . ,T$ \textbf{do}
		\STATE {\bfseries } $(c(t),r(t))$ is drawn according to distribution $D_{c,r}$
\STATE {\bfseries }
  \begin{equation}
       c(t) = 
        \begin{cases}
             \hat{c}:=\psi(c(t)) & \text{with probability $p_{\psi}$} \\
               c(t) & \text{otherwise}
        \end{cases}
    \end{equation}
\STATE{\bfseries} The player chooses an arm $k(t) = \pi(c(t))$
				\STATE {\bfseries } The reward $r_{k(t)}$ is revealed
		\STATE {\bfseries } The player updates its policy $\pi$
 \STATE {\bfseries } \quad\textbf{End do}
	\end{algorithmic}
\end{algorithm}

\begin{definition}[Cumulative regret]
The regret of a CBCC-solving algorithm accumulated during $T$ iterations is given as:
\begin{equation*}
R(T) = \max_{\pi \in \Pi } \sum ^{T}_{t=1} r_{\pi(\textbf{c}(t))}(t) - \sum^{T}_{t=1} r_{k(t)}(t).
\end{equation*}
\end{definition}

\subsubsection{Corruption is not detectable from the context}
\label{subsec:UCB}

\begin{algorithm}[ht]
 \caption{Contextual TS with Corrupted Context (CTSCC)}
\label{alg:TSRC}
\begin{algorithmic}[1]
 \STATE {\bfseries }\textbf{Require:} The initial values $S_k(0)$ and $F_k(0)$ of the Beta distribution parameters, and list of $\alpha_1,,\alpha_2$ policies. 
 \STATE {\bfseries }\textbf{Initialize:} $\forall k \in \{1,...,K\}, B_k=I_d$, $\hat{\mu_k}= 0_d, g_k = 0_d$, and $\forall k \in \{1,...,K\}$, $n_k(0)=0$.
 \STATE {\bfseries }\textbf{Foreach} $t = 1, 2, . . . ,T$ \textbf{do}
  \STATE {\bfseries }\quad\textbf{Foreach} Policy $\alpha= 1,2$ \textbf{do}
  \STATE {\bfseries }\quad \quad Sample $\tilde{\mu_{\alpha}}$ from $N(\hat{\mu_{\alpha}}, v^2 B_{\alpha}^{-1})$ distribution.
  \STATE {\bfseries } \quad\textbf{End do}
   \STATE {\bfseries }\quad Select Policy  $\alpha(t)=\underset{k\subset \{1,...,K\} }{argmax} c(t)^\top \tilde{\mu_{\alpha}}$
 \STATE {\bfseries }\quad\textbf{Foreach} arm $k= 1,...,K$ \textbf{do}
 
 \STATE {\bfseries }\quad \quad Sample $\tilde{\mu_k}$ from $N(\hat{\mu_k}, v^2 B_k^{-1})$ distribution.
\STATE {\bfseries } \quad\quad Sample $\theta_k$ from $ Beta(S_k(t), F_k(t))$ distribution
 \STATE {\bfseries } \quad\textbf{End do}
 \STATE {\bfseries }\quad Select arm  $k(t)=\underset{k\subset \{1,...,K\} }{argmax} \alpha(t) [c(t)^\top \tilde{\mu_k}]$ + $(1- \alpha(t))$ $[c(t)^\top \tilde{\mu_k}]$
 \STATE {\bfseries }\quad Observe $r_{k}(t)$
 \STATE {\bfseries }\quad\textbf{Foreach} arm $k= 1,...,K$ \textbf{do}
\STATE \quad\quad$B_k= B_{k}+ c(t)c(t)^{T} $, $g_k = g_k + c(t)r_{k}(t)$, $\hat{\mu_k} = B_k^{-1} g_k$
\STATE \quad\quad $n_k(t)=n_k(t-1)+1$,  $S_k(t)=S_k(0)+r_k(t-1)$, $F_k(t)=F_k(0)+n_k(t-1)-r_k(t-1)$
 \STATE {\bfseries } \quad\textbf{End do}
  \STATE {\bfseries }\quad\quad $S_{\alpha}(t)=S_{\alpha}(0)+r_{\alpha}(t-1)$, $F_{\alpha}=F_{\alpha}(0)+n_{\alpha}(t-1)-r_{\alpha}(t-1)$
\STATE {\bfseries }\quad\textbf{End if}
 \STATE {\bfseries }\textbf{End do}
 \end{algorithmic}
\end{algorithm}

We now propose a method for solving the CBCC problem, called {\em Thompson Sampling with corrupted Context (TSRC)}, and summarize it in Algorithm \ref{alg:TSRC} (see section \ref{background} for background on Thompson Sampling).

Let $n_k(t)$ be the number of times the $k$-th arm has been selected so far, and let $r_k(t)$ be the reward associated with the arm $k$ at time $t$.

We assume that reward $r_{k}(t)$ for choosing arm $k$ at time $t$ follows a parametric likelihood function $Pr(r(t)|\tilde{\mu_k})$, and that
the posterior distribution at time $t + 1$, $Pr(\tilde{\mu}|r(t)) \propto Pr(r(t)|\tilde{\mu}) Pr(\tilde{\mu})$:

- When it observes the true context the posterior is given by a multivariate Gaussian distribution $\mathcal{N}(\hat{\mu_k}(t+1)$, $v^2 B_k(t + 1)^{-1})$, where
$B_k(t)= I_d + \sum^{t-1}_{\tau=1} c(\tau) c(\tau)^\top$
with $d$ the size of the context vectors $c$, $v= R \sqrt{\frac{24}{\epsilon} d ln(\frac{1}{\gamma})}$ with $R>0$,  $\epsilon \in ]0,1]$, $\gamma \in ]0,1]$ constants, and $\hat{\mu}_k=B_k(t)^{-1} (\sum^{t-1}_{\tau=1} c(\tau) r(\tau))$. 

- When it observes corrupted context, we assume that the problem is a classical multi-armed bandit without context, so the posterior is a beta distribution. 

At each time point $t$, and for each arm, we sample a $d$-dimensional $\tilde{\mu_k}$ from $\mathcal{N}(\hat{\mu_k}(t)$, $ v^2B_k(t)^{-1})$ and sample the parameter $\theta_k$ from the corresponding $Beta$ distribution, then choose an arm maximizing $\alpha [c(t)^\top \tilde{\mu_k}]$ + $(1- \alpha)$ (step 9 in the algorithm).

The $\alpha$ parameter is binary and  allows us to decide which policy to trust most. The algorithm is solving a bandit problem where the $\alpha_1$ and $\alpha_0$ the two policies (the contextual and the classical) bandit are the arms of the bandit problem. The algorithm is using the Thompson sampling strategy, so it samples $\theta_\alpha$ from $ Beta(S_\alpha(t), F_\alpha(t))$ distribution.

Finally, the reward $r_k(t)$ for choosing an arm $k$ is observed, and relevant parameters are updated.
We update the values of those parameters, $S_k$ and $F_k$ (steps 16 and 17), to represent the current total number of successes and failures, respectively.

\subsection{Theoretical analysis}
We study in the following both the regret upper bound for the TSCC and the regret lower bound for the CBCC.

\subsubsection{Regret upper bound for the TSCC }
We derive here an upper bound on the regret of the policy computed by the Thompson Sampling with Corrupted Context, or TSCC, algorithm we just presented. We study this regret for two types of optimal policies. The optimal policy described in $Definition$ $\ref{defOP1}$ assumes having the optimal parameters for both the CMAB setting.
\begin{definition}[Optimal Policy]
\label{defOP1}
 The optimal policy for solving the CBCC is as follow:
\begin{align*}
k(t) =  \underset{k\subset \{1,...,K\} }{argmax} [ \alpha^*(t) \mu_{k}^*(t)^\top c(t)+(1-\alpha^*(t)) \theta^*(t)_k]
\end{align*}

where $\alpha^*$, $\mu_{k}^*$ and $\theta^*$ are respectively the optimal weight parameter, the optimal mean vector and the optimal mean value.
\end{definition}

\begin{thm}\label{theorem2}
Using definition $\ref{defOP1}$ of the optimal policy, with probability $1-\gamma$, where $0 < \gamma < 1$, the regret $R(T)$ accumulated by the $TSCC$ algorithm in $T$ iterations is upper-bounded by
\begin{eqnarray*}
R(T) \leq  \frac{d\gamma}{\epsilon} \sqrt{(T-\tau) ^{\epsilon+1}} (\ln (T-\tau ) d) \ln \frac{1}{\gamma}+l
\end{eqnarray*}
\end{thm}

with \quad $\mathbb{E}(l) \leq (\sum_{k\in S}\frac{1}{\Delta_k^2})^2 \ln\ T$

The Theorem \ref{theorem2} is showing that the upper bound of the TS for CBCC, is the combination of the two regret got in the Classical bandit setting and the contextual bandit setting.  
\begin{proof}
Our upper bound is based on two key results presented in the following lemmas.

\begin{lem}\label{fact:upperCBTS} \cite{AgrawalG13}
With probability $1-\gamma$, where $0 < \gamma < 1$, the upper bound on the expected regret R(T) for the CTS (Algorithm 3) in the contextual bandit problem  with $K$ arms and $d$ features (context size) is given as follows:
\begin{eqnarray*}
R(T)<\frac{d\gamma}{\epsilon} \sqrt{T^{\epsilon+1}} (ln (T) d) ln \frac{1}{\gamma}.
\end{eqnarray*}
\end{lem}

\begin{lem}\label{fact:uppercom} \cite{AgrawalG13}
The upper bound on the regret $R(T)$ for the TS (Algorithm 1) in the bandit problem with the set of $S = \{1,...,K\}$ arms is the following:
\begin{eqnarray*}
E[R(T)]<(\sum_{k\in S}\frac{1}{\Delta_k^2})^2 \ln\ T
\end{eqnarray*}
where  $\Delta_k=\mu^*- \mu_k$, where $\mu^*, \mu_k$ are the  expected rewards of the optimal arm   and of the arm   $k \in S$, respectively.
\end{lem}

We will split the  total  regret into two parts,
\begin{eqnarray*}
R(T)=R(\tau_1)+R(\tau_2),
\end{eqnarray*}
where $\tau_1$ is the number of time points within the first $T$ iterations at  which a sub-optimal policy was selected, and $\tau_2 $ is, vice versa, the number of times points when the best policy was used; clearly,   $T=\tau_1+\tau_2$.
Note that
\begin{eqnarray*}
	\tau_1= R_{bandit}(T),
\end{eqnarray*}
where $R_{bandit}(T)$ is the regret of the bandit accumulated in the first part of the algorithm when looking for the optimal policy.  Then
\begin{eqnarray*}
	R(T) = R_{bandit}(T) + R_{L_2}(\tau_2),
\end{eqnarray*}
where $R_{L_2}(\tau_2)$ is the regret accumulated by the second level bandit algorithm, because while the algorithm is playing $\tau_2$ times with the optimal policy, it is not making mistakes due  to  policy selection.

So using lemma \ref{fact:upperCBTS} and lemma \ref{fact:uppercom}, we respectively upper bound the first term and the second term of the equation, which gives us the final results.
\end{proof}

\subsubsection{Regret lower bound for the CBCC}
We derive here a lower bound on the regret for the CBCC. 

\begin{thm}\label{fact:lowCBCC} 
For any algorithm solving the CBCC problem with context size $d$, with $(1-p_{\psi})$ and $0 \leq p_{\psi} \leq 1$ there exists a constant $\gamma > 0$, such that the lower bound of the expected regret accumulated by the algorithm over $T$ iterations is lower-bounded as follows:
\begin{eqnarray*}
E[R(T)]>\gamma \sqrt{Td}.
\end{eqnarray*}
\end{thm}

where $p_{\psi}$ is the probability that the context is corrupted by an unknown function $\psi$.
\begin{proof}
Our lower bound is based on the key result presented in the following lemmas.

\begin{lem}\label{fact:lowCB} \cite{ChuLRS11}
For any algorithm solving the {\em contextual bandit} problem with the context size $d$, there exists a constant $\gamma > 0$ such that the lower bound of the expected regret accumulated by the algorithm over $T$ iterations is lower-bounded as follows:
\begin{eqnarray*}
E[R(T)]>\gamma \sqrt{Td}.
\end{eqnarray*}
\end{lem}
This theorem shows that in the best case scenario any algorithm solving the CBCC is going to have the same error lower bound as classical contextual bandit setting. 

The proof is straightforward: the regret at time $T$ is as follows: 

\begin{align*}
R(T) =  R(T_c)+R(T_{nc})
\end{align*}

with $T_c$ the time that the agent got corrupted context and $T_{nc}$ the time that the agent got uncorrected context.

From the problem definition we have that with probability $(1-p_{\psi})$ we have $R(T) = R(T_{nc})$ which is the best case scenario in this setting. So, using lemma \ref{fact:lowCB} to lower bound $R(T_{nc})$ we get our final result.
\end{proof}

\section{Empirical Evaluation}
\label{sec:experimental}
Empirical evaluation of the proposed methods was based on four datasets from the UCI Machine Learning Repository \footnote{https://archive.ics.uci.edu/ml/datasets.html}: Covertype, CNAE-9, Internet Advertisements and Poker Hand (for details of each dataset, see Table \ref{table:Synthetic}).

\begin{table}[t]
	\centering
	\caption{Datasets}
	\resizebox{0.6\columnwidth}{!}{
		\begin{tabular}{ l | c | r | r }
			UCI Datasets                  & Instances    & Features   & Classes \\ \hline
			Covertype                & 581 012      & 95           &  7\\
			CNAE-9                   & 1080       & 857          &  9\\
			Internet Advertisements  & 3279         & 1558         &  2\\
			Poker Hand               & 1 025 010    & 11         &  9\\

		\end{tabular}
	}
	\label{table:Synthetic}
\end{table}

\begin{table*}[tbh]
	\centering
	\caption{Stationary Environment}
	\resizebox{0.95\columnwidth}{!}{
		\begin{tabular}{ l | l | l | l | l | l | l}
			                         &  MAB               &  NSMAB              & CMAB                             & TSCC         \\ \hline
			Datasets \\ \hline
			Covertype                & $70.54\pm 0.30$    &  $ 75.27\pm 2.49$   & $ 65.54\pm 4.57$  &  $\textbf{63.44}\pm\textbf{3.75}$ \\
			CNAE-9 	                 &  $79.85 \pm 0.35$  &   $82.01\pm 1.39$	& $ 73.47\pm 3.55$ &  $\textbf{70.47}\pm \textbf{1.93}$\\
			Internet Advertisements	 &  $19.22\pm0.20$	  &   $21.33\pm 1.38$	& $ 16.21\pm 2.54$  &  $\textbf{15.21}\pm\textbf{1.10}$  \\
			Poker Hand 	             &  $62.29\pm 0.21$   &   $ 68.57 \pm 1.17$	& $ 58.82\pm 3.89$  &  $\textbf{57.20}\pm\textbf{0.91}$  \\
			\hline
		\end{tabular}
	}
	\label{table:AccuSyn}
\end{table*}

\begin{figure*}[tbh]
	\begin{multicols}{2}
		\includegraphics[height=1.5 in,width=0.99\linewidth]{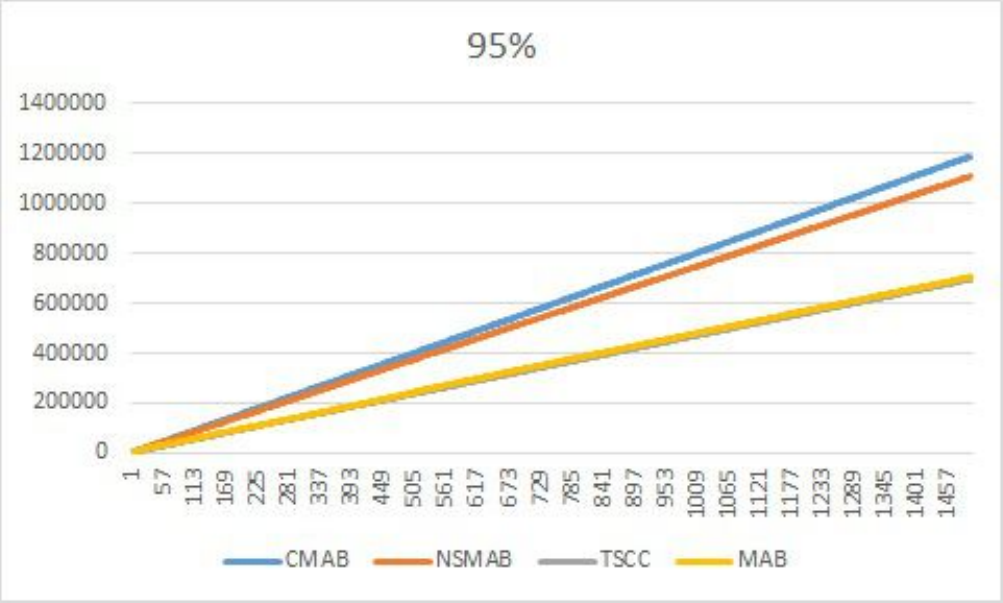}\par
\vspace{0.2in}		
		\includegraphics[height=1.5 in,width=0.99\linewidth]{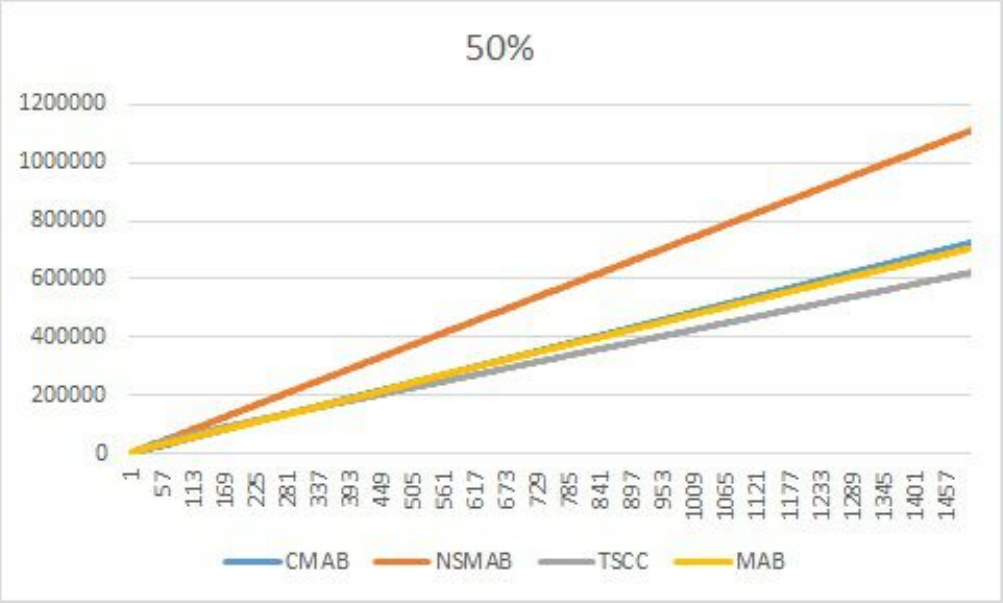}\par
\vspace{0.2in}	
		\includegraphics[height=1.5 in,width=0.99\linewidth]{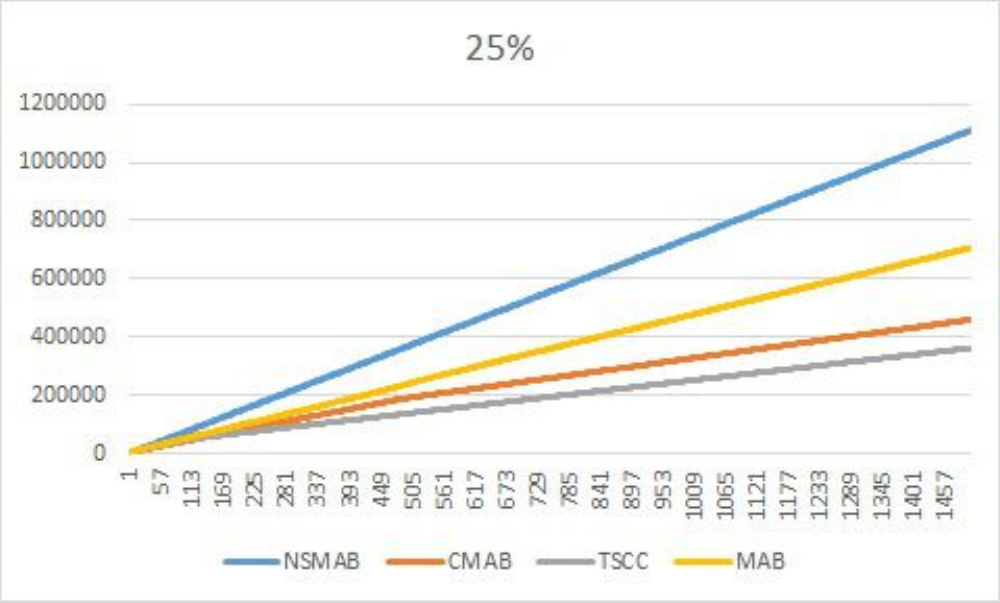}\par
	\end{multicols}
	\caption{ Covertype dataset}\label{fig:incrst}
\end{figure*}

In order to simulate a data stream, we draw samples from the dataset sequentially, starting from the beginning each time we draw the last sample. At each round, the algorithm receives the reward 1 if the instance is classified correctly, and 0 otherwise.  We compute the total number of classification errors as a performance metric.

We compare our algorithms with the following competing methods:
\begin{itemize}
	\item {\em Multi-Arm Bandit (MAB)}: this is the standard Thomspon Sampling approach to (non-contextual) multi-arm bandit setting.
		\item {\em Non-Stationary Multi-Arm Bandit (NSMAB)}: we have use here WUCB proposed in \cite{garivier45} approach as a baseline to (non-contextual) multi-arm bandit setting.
	\item {\em Contextual Bandit (CMAB)}: this algorithm uses the contextual Thomspon Sampling (CTS).
\end{itemize}

We ran the above algorithms and our proposed TSCC method for different corrupted context subset sizes, such as 5\%, 25\%, 75\% and 95\% of corrupted context. To do that we have added some contexts with random values in the set that we are sampling from. 

\subsection{Analysis in the four datasets}
Table ~\ref{table:AccuSyn} summarizes our results for the CBCC setting; it represents the average classification error, i.e. the misclassification error, computed as the total number of misclassified samples over the number of iterations. This average errors for each algorithm were computed using 10 cyclical iterations over each dataset, and over the four different corrupted levels mentioned above.

Overall, based on their mean error rate, our TSCC approach shows superior performance (shown in bold in Table ~\ref{table:AccuSyn}) when compared to the rest the algorithms, underscoring the importance of combining the multi-armed bandit with the contextual bandit as we did in this setting.

\subsubsection{Detailed analysis on Covertype dataset}
Figure \ref{fig:incrst} provides a more detailed evaluation of the algorithms for different levels of corrupted contexts, on a specific dataset. We observe that :

{\bf 95 \% corrupted:} {\em MAB} has the lowest error out of all methods, followed tightly by {\em NSMAB}, suggesting that, ignoring the context in MAB may still be a better approach than doing the contextual bandit when the number of corrupted context is high. we also observe that {\em MAB} has a lower error than the {\em NSMAB}, which means that a stationery algorithm is better to handle the corrupted context compered with a non-stationary approach.

{\bf 50 \% corrupted:} {\em TSCC} has the lowest error. However, at this corruption level we can see that CMAB and MAB have the same level of accuracy. 

{\bf 25 \% corrupted:} TSCC has the lowest error, followed by the {\em CMAB}, which implies that, which implies that at this corruption level, the need for the MAB strategy is very low.

\section{Conclusions}
\label{sec:Conclusion}
We have introduced a new formulation of MAB, motivated by several real-world applications including medical diagnosis and recommender system. In this setting, which we refer to as {\em contextual bandit with corrupted context (CBCC)}, a set of features, or a context that is used to describe the current state of world could corrupted. So, the agent can not always trust the observed context, and thus needs to combine contextual bandit approach with a classical multi-armed bandit mechanism, in order to comput the expectation of the arms. We proposed novel algorithm based on Thompson Sampling for solving the CBCC problem. Furthermore, we derived upper bound for the proposed algorithm. Empirical evaluation on several datasets demonstrates advantages of the proposed approach.



\bibliographystyle{unsrt} 
\bibliography{example_paper}

\end{document}